\documentclass{ecai} 

\usepackage{latexsym}
\usepackage{amssymb}
\usepackage{amsmath}
\usepackage{amsthm}
\usepackage{booktabs}
\usepackage{enumitem}
\usepackage{graphicx}
\usepackage{color}
\usepackage{multirow}
\usepackage{bm}

\newcommand{\BibTeX}{B\kern-.05em{\sc i\kern-.025em b}\kern-.08em\TeX}

\begin{document}


\begin{frontmatter}



\title{Keyword-Centric Prompting for One-Shot Event Detection with Self-Generated Rationale Enhancements}


\author[A]{\fnms{Ziheng}~\snm{Li}}
\author[A]{\fnms{Zhi-Hong}~\snm{Deng}\thanks{Corresponding Author. Email: zhdeng@pku.edu.cn}}

\address[A]{State Key Laboratory of General Artificial Intelligence, School of Intelligence Science and Technology, Peking University}


\begin{abstract}
Although the LLM-based in-context learning (ICL) paradigm has demonstrated considerable success across various natural language processing tasks, it encounters challenges in event detection.
This is because LLMs lack an accurate understanding of event triggers and tend to make over-interpretation, which cannot be effectively corrected through in-context examples alone.
In this paper, we focus on the most challenging one-shot setting and propose KeyCP++, a keyword-centric chain-of-thought prompting approach.
KeyCP++ addresses the weaknesses of conventional ICL by automatically annotating the logical gaps between input text and detection results for the demonstrations.
Specifically, to generate in-depth and meaningful rationale, KeyCP++ constructs a trigger discrimination prompting template.
It incorporates the exemplary triggers (a.k.a keywords) into the prompt as the anchor to simply trigger profiling, let LLM propose candidate triggers, and justify each candidate.
These propose-and-judge rationales help LLMs mitigate over-reliance on the keywords and promote detection rule learning.
Extensive experiments demonstrate the effectiveness of our approach, showcasing significant advancements in one-shot event detection.
\end{abstract}

\end{frontmatter}


\section{Introduction}
\begin{figure}[t]
    \centering
    \includegraphics[width=0.8\linewidth]{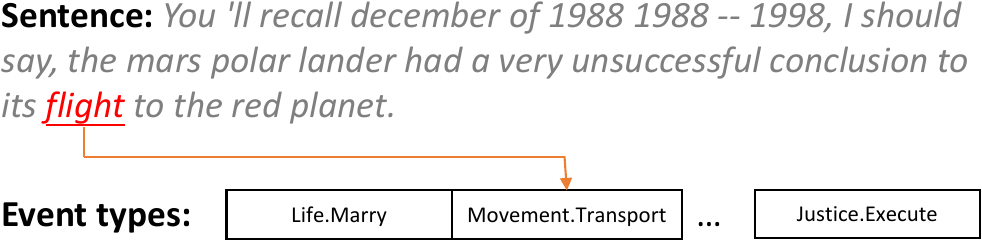}
    \caption{An event detection example. The sentence mentions a \textit{Movement.Transport} event.}
    \label{fig:task}
\end{figure}

Event Detection (ED) is the task of identifying event triggers of predefined types within a given text. For example, in the sentence shown in Figure~\ref{fig:task}, there is a \textit{Movement.Transport} event whose trigger is "\textit{flight}". ED plays a fundamental role in various NLP tasks, such as knowledge graph construction~\citep{zhang2020aser} and question answering~\citep{han2021ester,li-etal-2020-gaia}.

Traditional ED approaches~\citep{nguyen_joint_2016,wadden_entity_2019,liu_event_2020, lin_joint_2020, pouran_ben_veyseh_unleash_2021} heavily rely on supervised fine-tuning and necessitate extensive annotated training data. This paradigm faces great challenges for real-world deployment due to the emergence of new event types and the high cost associated with data annotation. The advancements of large language models (LLM) like GPT-4 and DeepSeek~\citep{deepseekai2025deepseekr1incentivizingreasoningcapability} introduce in-context learning (ICL)~\citep{brown_language_2020,wei_chain--thought_2022,kojima_large_2022} as a promising alternative solution for low-resource scenarios. Leveraging the vast general knowledge and instruction following ability acquired during pre-training, LLMs demonstrate innate proficiency as few-shot learners.

However, existing ICL approaches obtain poor performance when directly applied to the event detection task~\citep{wei_zero-shot_2023,gao_exploring_2023,guo_retrieval-augmented_2024}, showing little advantage compared with conventional supervised fine-tuning approaches.
Through in-depth analysis, we attribute the failure to two main reasons: 1) although LLMs may grasp the concept of target events, they lack an accurate understanding of triggers; 2) the in-context examples alone are insufficient for teaching LLMs the concept of triggers. Consequently, conventional ICL methods tend to miss obvious triggers or make over-interpretations.

\begin{figure}[t]
    \centering
    \includegraphics[width=0.8\linewidth]{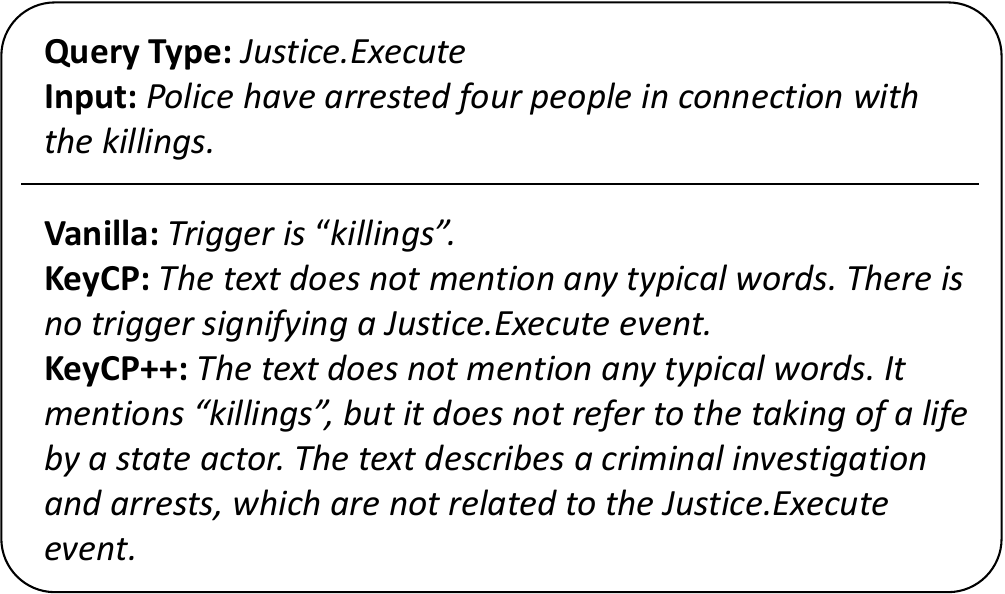}
    \caption{Example for different prompting strategies. Vanilla prompting misidentifies the non-execution killing as the trigger. KeyCP obtains the right answer because "killing" is not a usual expression of execution. KeyCP++ additionally takes "killing" into consideration and conducts an explicit definition check.}
    \label{fig:contrast}
\end{figure}

Inspired by chain-of-thought (CoT) prompting~\citep{wei_chain--thought_2022}, we aim to prompt LLMs to generate a reasoning process before arriving at the final answer to address the aforementioned weaknesses.
However, CoT prompting typically relies on curated rationale annotations to activate the model’s reasoning capabilities. This reliance poses scalability challenges, as obtaining high-quality annotations from domain experts is costly and impractical—especially given the continuous emergence of new event types.
A more scalable alternative is to enable LLMs to automatically generate rationales for demonstration examples.
The primary challenge in rationale generation lies in achieving logical richness. We observe that when prompted directly to explain an example, LLMs tend to reproduce surface-level definitions without meaningful interpretation.

To address this, we propose a novel keyword-centric rationale-enhanced prompting framework KeyCP++, which can automatically generate helpful rationales.
KeyCP++ is built on a base prompting framework KeyCP, which leverages keywords to steer the LLM output and provides KeyCP++ with a logically rich topic to generate rationales.
The utilization of keywords is inspired by previous supervised fine-tuning works~\citep{hsu_degree_2022,zhao_demosg_2023}.
Here the keywords refer to exemplary triggers or other words highly related to the target event, deduced from the definition.
These keywords can be either handcrafted or automatically generated.
A critical function of KeyCP is to align the LLM's trigger profile with these keywords.
To achieve this, we employ keywords to supplement event definitions and insert the keyword detection results into the prompt.
This approach forces the LLM to focus more on event-related text and reduce over-interpretation.

KeyCP++ inserts rationale into the KeyCP prompting template to provide further guidance in learning from the in-context examples and prevent LLMs from over-relying on the keywords.
To this end, we introduce a proposal-judgment workflow.
Unlike KeyCP which uses a fixed set of keywords, KeyCP++ allows LLMs to propose trigger candidates at the beginning of the generation as a supplement to the keywords.
Subsequently, LLMs will generate rationales that judge whether each keyword and proposed candidate conform to the event definition.
We devise an automatic procedure to annotate the proposals and judgments of the in-context examples, which are then incorporated into the prompt to guide the generation during inference.
Compared with KeyCP, KeyCP++ offers more flexibility because the detection is not limited to predefined keywords, and the rationales help LLMs learn the internal process of identifying triggers.
To demonstrate the generality and robustness of our method, we evaluate our approaches using LLaMA2-13B~\citep{touvron_llama_2023}, Mistral-7B~\citep{jiang_mistral_2023}, GPT3.5, and DeepSeek-V3.
Our results show that in one-shot event detection scenarios, KeyCP++ significantly outperforms prior ICL and supervised fine-tuning SOTA.

Our contributions are summarized as follows:
\begin{itemize}
    \item We introduce a strong baseline KeyCP which significantly mitigates the trigger profiling problem in ICL.
    \item We propose a novel rationale-enhanced framework KeyCP++ that further improves the flexibility and learning ability of KeyCP. To the best of our knowledge, we are the first to present an effective chain-of-thought prompting paradigm for event detection.
    \item We substantially improve the performance of in-context learning in event detection as demonstrated in the extensive experiments on ACE2005~\citep{ace2005} and WikiEvents~\citep{li_document-level_2021}.
\end{itemize}

\begin{figure*}[t]
    \centering
    \includegraphics[width=0.95\linewidth]{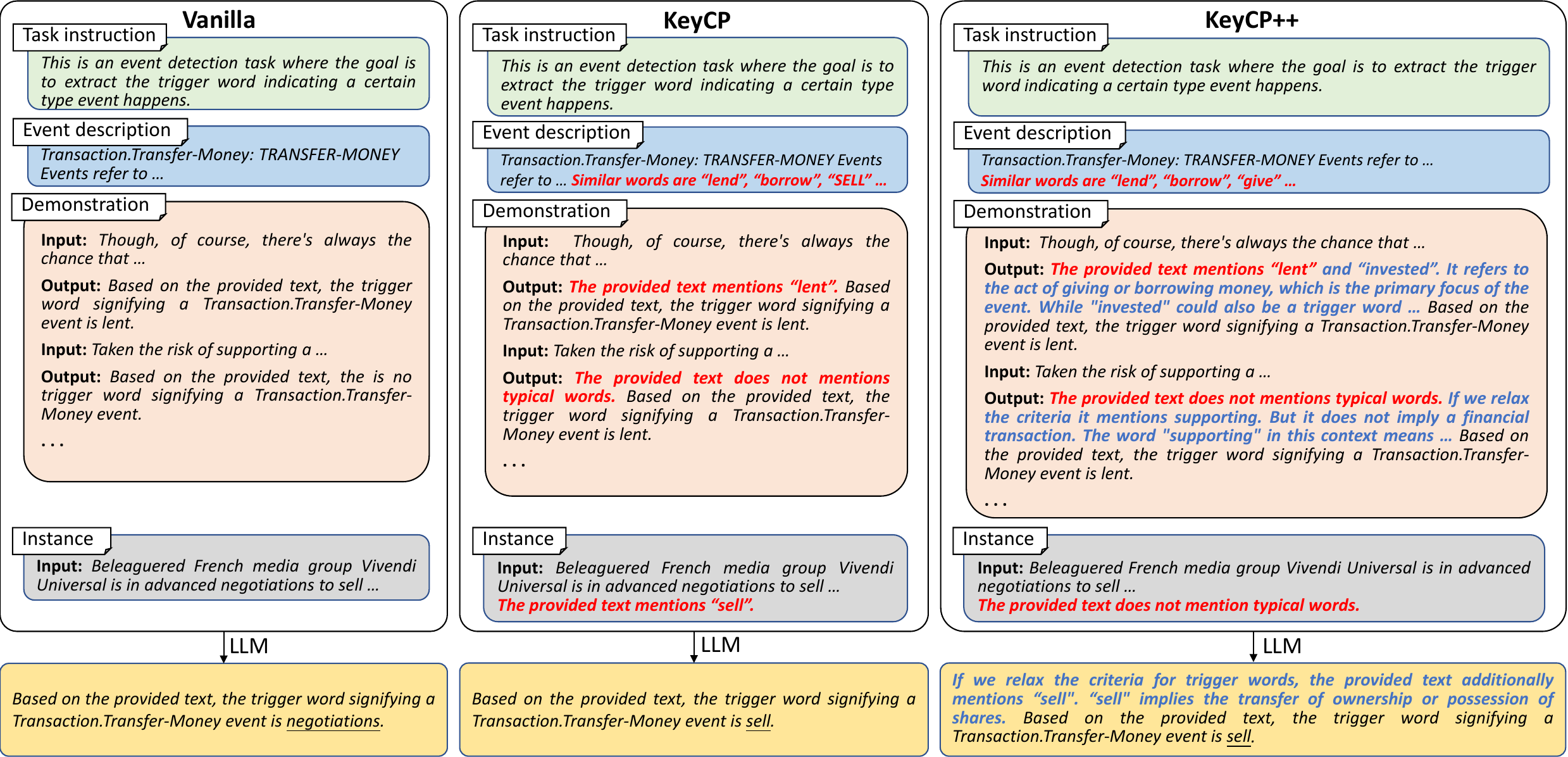}
    \caption{Overview of the vanilla, KeyCP, and KeyCP++ prompting. A prompt comprises task instruction, event description, demonstration, and instance. We parse the trigger (underlined) from the generation. Compared with vanilla prompting, KeyCP adds keyword list and detection to event description, demonstration and instance respectively (red). On the basis of KeyCP, KeyCP++ adds proposal-judgment rationale (blue) for each example. The complete prompting examples can be found in the Appendix~\ref{apsec:example}.}
    \label{fig:overview}
\end{figure*}

\section{Related Work}
\subsection{Event Detection}
As an important natural language processing task, event detection has been studied for decades.
It often appears as a sub-task in the event extraction literature.
Most existing works train their models on annotated datasets in a supervised learning manner. Early works usually treat event detection as a token classification task~\citep{nguyen_joint_2016,yang_exploring_2019,wadden_entity_2019, lin_joint_2020, pouran_ben_veyseh_unleash_2021}.
Some researchers augment the original sentence with a designed QA template to enhance classification performance~\citep{li_event_2020,liu_event_2020,du_event_2021,huang_iteratively_2023}. Recently, many works have formulated event detection as a text generation task to leverage the capabilities of powerful pre-trained generative language models.
\citet{lu_text2event_2021} introduces a linearized format for the event structure so that the training target can be transformed into a text sequence. The application of prompting techniques~\citep{brown_language_2020} further narrows the gap between event detection and language model pre-training \cite{hsu_degree_2022, liu_dynamic_2022, dafs,zhao_demosg_2023}.
These works design type-specific templates incorporating the event definition and structure information and let the language model fill the trigger placeholders. Benefiting from the pre-trained models' knowledge and manual templates, template-based methods exhibit better performance in low-source scenarios.

\subsection{In-Context Learning for Event Detection}
In-context learning (ICL) is a new few-shot learning paradigm~\citep{brown_language_2020} wherein LLMs learn a task from the demonstration formed by a few examples rather than gradient updates. The performance of ICL strongly depends on the prompt design. Researchers have found ICL with simple input-output pairs struggles on complex tasks requiring commonsense and reasoning even when using the most powerful models.
To further facilitate LLM's few-shot ability, \citet{wei_chain--thought_2022} proposed chain-of-thought (CoT) prompting where they insert a rationale before each example's answer.
These rationale-augmented demonstrations will guide the LLM to output a series of intermediate reasoning steps.
Many works have found that CoT significantly outperforms the standard ICL prompting~\citep{wei_chain--thought_2022,zhang_what_2025,ma_non-myopic_2025,wu_inference_2024, wang_strategic_2024-1}.
The advancements of in-context learning inspire researchers to explore fine-tuning-free approaches for event detection~\citep{xu_large_2024,wang_code4struct_2023,chen2024large}.
\citet{gao_exploring_2023} utilizes ChatGPT~\citep{openai_gpt-4_2023} to generate JSON format event structure by prompting with simple input-output demonstration pairs.
\citet{guo_retrieval-augmented_2024} formalizes event extraction as Python code completion, where each event type is represented by a well-documented Python class.
However, their performance is non-competitive with the fine-tuning-based approaches.
\citet{pang_guideline_2023} proposes to add extraction guidelines in the prompt where the guidelines are generated from the wrong predictions by LLMs. But their approach is limited to trigger classification, leaving trigger identification unsolved.

\section{Methodology}
\subsection{Formulation of One-Shot Event Detection}
For a predefined set of event types $T=\{t_i\}_{i=1:K}$, given a query sentence $s$ along with a query type $t\in T$, the event detection task requires models to determine whether $s$ contains one or more events belonging to type $t$ and identify their trigger words that signify the occurrence of events. Models are required to learn the task from a training set containing one labeled example for each event type, supplemented by some high-level description such as event definitions $D=\{d_1,d_2,\cdots,d_K\}$ and keywords $\mathcal{W}=\{W_1,W_2,\cdots,W_K\}$.

\subsection{Keyword-Centric Prompting}
KeyCP is a prompt-based method. We formulate the event detection problem as a text generation task. Unlike previous ICL works, we query one event type per forward propagation similar to supervised fine-tuning methods~\citep{hsu_degree_2022,liu_dynamic_2022,zhao_demosg_2023}, because concatenating all event types in one prompt will result in a very long input which exceeds the maximum context length of many LLMs. Given a query instance $x$ and a query type $t$, we detect the event mention $f_t(x)$ as follows:
\begin{equation}
    f_t(x)=h(LLM(g(x,d_t,W_t,e_t,\bar e_t^1,\cdots,\bar e_t^S))),
\end{equation}
where $e_t$ represents the positive example corresponding to the query type, and $\bar e_t^i$ denotes the negative example sampled from other types. In our experiments, we set the negative sampling size $S=5$. $g$ is the prompting function that integrates the task instruction, event description, training examples, and query instance as the input of the LLM. After generation, we parse the answer from the output of LLM by a pattern-matching algorithm $h$.

Vanilla prompting methods simply concatenate the event definitions and the input-output pairs as illustrated in Figure~\ref{fig:overview} (left). However, it is difficult for LLMs to apply the extraction criteria implicated in the event definition, or capture the intricate relationship between a noisy sentence and a single word. Consequently, LLMs largely depend on their prior knowledge and preference alignment to make predictions. For example, in Figure~\ref{fig:overview} (left), both "sell" and "negotiations" are related to the \textit{Transfer-Money} event, but the proper trigger is "sell" rather than "negotiations" in this task. Vanilla prompting is not able to distinguish them and may output the wrong one.

To simplify task learning, we propose directly informing LLMs which words are more likely to be triggers. We generate a set of keywords for each event type according to the event definition by GPT3.5.\footnote{The quality of keywords may impact the detection results, but the acquiring of keywords is not the focus of this paper. One can use handcrafted keywords for stable and better performance.} The generation details can be found in the Appendix~\ref{apsec:keywords}. The KeyCP prompting template is shown in Figure~\ref{fig:overview} (middle). We incorporate these keywords into the event description and the demonstration. To ensure LLMs can always notice the keywords and avoid fabrication, we perform a string matching for the query instance and append the matching results to the prompt. Intuitively, KeyCP uses the keywords as the anchor and lets LLMs make further judgments based on them. In contrast to the marginal benefits of leveraging keywords in fine-tuning approaches\citep{hsu_degree_2022}, KeyCP demonstrates substantial performance enhancements of 30\% at most (shown in Section~\ref{sec:results}).

\begin{figure*}[t]
    \centering
    \includegraphics[width=0.99\linewidth]{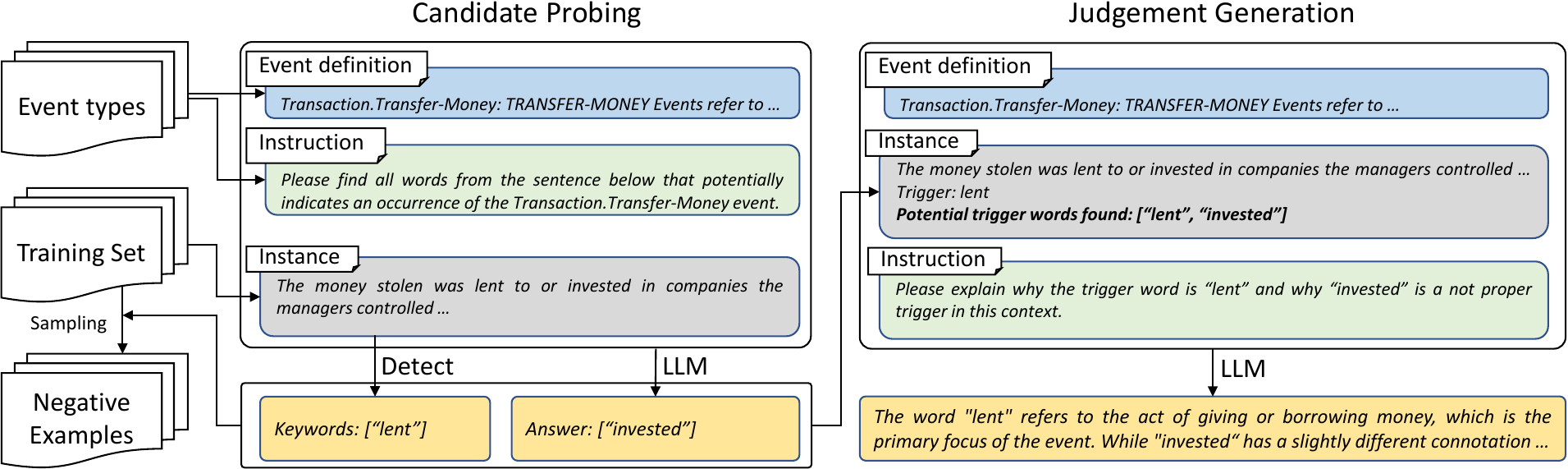}
    \caption{Illustration of the rationale generation. First, we feed every example-type pair into the LLM to probe candidate triggers (left). Then, we perform negative example sampling prioritizing examples holding more candidates (left bottom). Finally, we prompt the LLM to discriminate candidates from the golden label (right).}
    \label{fig:rationale}
\end{figure*}

\subsection{Rationale Enhancement}
Although introducing keywords helps LLMs profile the triggers, it may lead to over-reliance on them. Besides, KeyCP provides little help in learning the detection rule. Therefore, we propose KeyCP++ to enhance LLMs' ability of trigger discrimination and contextual reasoning. KeyCP++ is inspired by chain-of-thought reasoning~\cite{wei_chain--thought_2022}. We add rationales to the demonstration to fill the blank of intermediate trigger detection steps as shown in Figure~\ref{fig:overview} (right). KeyCP++ builds on KeyCP's use of keywords to anchor trigger profiling and further encourages LLMs to explore non-keyword proposals. These generated proposals serve as the less reliable trigger candidates. We then prompt LLMs to thoroughly judge whether and how each candidate is involved in an event, so that improper candidates (both keywords and proposals) will be filtered out.

To annotate rationales for the examples, we devise an automatic rationale generation framework while we do not introduce any human supervision. The rationale generation framework consists of three parts: candidate probing, negative sampling, and judgment generation as illustrated in Figure~\ref{fig:rationale}.

\paragraph{Candidate Probing.}
There are two purposes of candidate probing. The first is to emulate the process of proposing non-keyword words. The second is to discover LLMs' inherent bias about event triggers for subsequent alignment. To achieve both purposes simultaneously, we instruct the LLM to perform a zero-shot detection without keywords for each training example, querying each event type. The generated proposals, along with the detected keywords, serve as the trigger candidates.

\paragraph{Negative Sampling.}
In the vanilla ICL method, negative examples are sampled uniformly from the training set. However, in KeyCP++, examples with more candidates contain more information. So, we sample negative examples for event type $t$ with the following probability distribution:
\begin{equation}
    P_t(x)=\frac{1}{Z}e^{|C_t(x)|/\tau},
\end{equation}
where $C_t$ denotes the candidate set of $x$ for type $t$, $\tau$ is the temperature controlling the concentration (set to $1$ in our experiments), and $Z$ is the normalizing factor.

\paragraph{Judgment Generation.}
For positive examples, given the probed candidates and golden trigger, we instruct the LLM to explain why the golden label is the most appropriate trigger and not the other candidates. By making these comparisons, the LLM will learn the characteristics of the correct trigger. Similarly, for negative examples, we instruct the LLM to explain why the text does not contain an event even if it mentions some plausible triggers. This procedure helps rectify any biased trigger profiling identified during candidate probing.

During inference, we assemble both the candidates and judgments in the demonstration. The LLM will imitate the demonstration to propose candidates and make judgments.

\setlength\tabcolsep{15pt}
\begin{table*}[t]
    \centering
    \caption{Performance comparison between KeyCP++, KeyCP, and vanilla prompting. We report averaged results and standard deviations across five different seeds and data splittings. Underlined data represents the best performance within a group.}
    \begin{tabular}{ccrrr}
        \toprule
        Method&Model&ACE05-E&ACE05-E\textsuperscript{+}&WikiEvents\\
        \midrule
        Vanilla & Mistral-7B& $20.9_{\pm0.9}$ & $24.0_{\pm2.2}$ & $18.7_{\pm0.7}$ \\
        KeyCP & Mistral-7B& $34.3_{\pm0.6}$ & $37.9_{\pm2.7}$ & $23.2_{\pm0.5}$ \\
        KeyCP++ & Mistral-7B& $\underline{38.7}_{\pm1.2}$ & $\underline{40.1}_{\pm0.8}$ & $\underline{24.4}_{\pm1.4}$ \\
        \midrule
        Vanilla & Llama2-13B& $8.9_{\pm0.6}$ & $9.0_{\pm0.3}$ & $4.5_{\pm0.0}$ \\
        KeyCP & Llama2-13B& $38.6_{\pm2.0}$ & $37.9_{\pm1.1}$ & $25.1_{\pm1.5}$ \\
        KeyCP++ & Llama2-13B& $\underline{45.1}_{\pm2.9}$ & $\underline{43.0}_{\pm1.9}$ & $\underline{29.3}_{\pm1.5}$ \\
        \midrule
        Vanilla & GPT3.5& $23.5_{\pm0.3}$ & $26.2_{\pm1.6}$ & $11.3_{\pm0.2}$ \\
        KeyCP & GPT3.5& $38.0_{\pm0.2}$ & $38.6_{\pm0.8}$ & $31.2_{\pm0.9}$ \\
        KeyCP++ & GPT3.5& $\underline{43.1}_{\pm0.1}$ & $\underline{45.0}_{\pm0.2}$ & $\underline{32.8}_{\pm0.5}$ \\
        \midrule
        Vanilla & DeepSeek-V3& $34.0_{\pm0.6}$ & $33.4_{\pm1.7}$ & $23.2_{\pm0.5}$ \\
        KeyCP & DeepSeek-V3& $49.6_{\pm0.5}$ & $44.3_{\pm1.8}$ & $36.2_{\pm0.3}$ \\
        KeyCP++ & DeepSeek-V3& $\underline{\bm{57.6}}_{\pm0.2}$ & $\underline{\bm{50.9}}_{\pm0.3}$ & $\underline{\bm {40.8}}_{\pm0.5}$ \\
        \bottomrule
    \end{tabular}
    \label{tab:main1}
\end{table*}

\section{Experiments}
In this section, we first describe our evaluation setup. Then we compare our approach with vanilla prompting, prior in-context learning works, and supervised fine-tuning methods.
\subsection{Datasets}
We evaluate on ACE2005~\citep{ace2005} and WikiEvents~\citep{li_document-level_2021}. The one-shot training sets are constructed by randomly sampling one instance for each event type from the full training set.

\textbf{ACE2005} is a widely used dataset for event extraction. It has 33 event types sourced from various media such as news, blogs, and broadcasts. Following the previous works, we conduct experiments on two pre-processing variants: ACE05-E~\citep{wadden_entity_2019} and ACE05-E\textsuperscript{+}~\citep{lin_joint_2020}. We leverage the event definitions provided in the official annotation guidelines.

\textbf{WikiEvents} is a recent dataset constructed from English Wikipedia. The original dataset has 67 event types in a three-level hierarchy. Considering the completeness of annotations, we only use the top 2 levels, resulting in 33 event types. We leverage the KAIROS ontology definitions as the event definitions.

\begin{table*}[t]
    \centering
    \caption{Performance comparison between KeyCP++ and previous in-context learning and fine-tuning event detection methods. For KeyCP++, we report the results using DeepSeek-V3.}
    \begin{tabular}{cccrrr}
        \toprule
        Method&Type&ACE05-E&ACE05-E\textsuperscript{+}&WikiEvents\\
        \midrule
        Text2Event & SFT & $13.7_{\pm1.9}$ & $17.3_{\pm1.8}$ & $13.2_{\pm1.9}$ \\
        UIE & SFT & \multicolumn{1}{c}{$38.1$} & \multicolumn{1}{c}{-} & \multicolumn{1}{c}{-} \\
        DEGREE & SFT & $44.8_{\pm2.0}$ & $43.8_{\pm2.7}$ & $28.2_{\pm3.9}$ \\
        ChatGPT & ICL & $43.8_{\pm0.7}$ & $42.7_{\pm1.4}$ & $36.3_{\pm0.4}$ \\
        CodeUIE & ICL & $45.1_{\pm1.4}$ & $45.6_{\pm2.2}$ & $39.0_{\pm2.1}$ \\
        KeyCP++ & ICL & $\underline{\bm{57.6}}_{\pm0.2}$ & $\underline{\bm{50.9}}_{\pm0.3}$ & $\underline{\bm {40.8}}_{\pm0.5}$ \\
        \bottomrule
    \end{tabular}
    \label{tab:main2}
\end{table*}
\setlength\tabcolsep{8pt}

\subsection{Implementation Details}
\paragraph{Metrics.} We use trigger classification F1 score in our experiments, following previous works~\citep{wadden_entity_2019,lin_joint_2020,hsu_degree_2022}. A prediction is considered correct if both the identified trigger offset and the classified event type match the golden standard.
\paragraph{Language model settings.} We evaluate our approach on the popular open-source language models LLaMA2-13B and Mistral-7B, and DeepSeek-V3. We also test the commercial LLM GPT3.5 (gpt-3.5-turbo-0125). During inference, we adopt the greedy decoding strategy. For candidate probing and judgment generation, we set the temperature to 0.9 and top-p to 0.6. We run LLaMA2-13B and Mistral-7B using 4 RTX A6000 GPUs and call API for GPT3.5 and DeepSeek-V3. LLaMA2-13B takes 1 hour to make inferences for each task and Mistral-7B takes 20 minutes.
\paragraph{Candidate probing.} For keyword detection, we use the NLTK lemmatizer~\citep{bird-2006-nltk} to obtain the stem of each word for matching. When making proposals, we repeat generation five times and take the output appearing more than three times.
\subsection{Baselines}
We first compare KeyCP++, KeyCP, and vanilla prompting to demonstrate the effectiveness of our chain-of-thought prompting framework. Additionally, we compare with previous in-context learning event detection works:
\begin{itemize}
    \item \textbf{ChatGPT}~\citep{gao_exploring_2023}, a preliminary work exploring prompting ChatGPT for event detection. It is similar to vanilla prompting but formalizes the event detection as a JSON writing task and etects all event types simultaneously.
    \item \textbf{CodeUIE}~\citep{guo_retrieval-augmented_2024} utilizes GPT3.5's coding ability and represents the event and trigger with Python class and object, formulating event detection as a code completion task.
\end{itemize}

We reproduce the results of ChatGPT CodeUIE using DeepSeek-V3~\citep{deepseekai2025deepseekr1incentivizingreasoningcapability}. We did not use DeepSeek-R1 because we found that R1's long-chain reasoning pattern does not help with event detection task. Besides, we also compare with previous supervised fine-tuning works. We choose three representative methods that perform well in one-shot settings:
\begin{itemize}
    \item \textbf{Text2Event}~\citep{lu_text2event_2021} converts the event record to a tree format that can be linearized as plain text so that they can formulate event detection as a seq2seq generation problem.
    \item \textbf{UIE}~\citep{lu_unified_2022} takes a similar strategy as Text2Event and improve low-resource performance through multi-task pre-training.
    \item \textbf{DEGREE}~\citep{hsu_degree_2022} incorporates additional event knowledge and employs manual prompts to further improve event detection.
\end{itemize}
We reimplement Text2Event and DEGREE using the largest language model reported in their respective papers. For UIE, we directly cite the results reported in their paper.

\subsection{Results}\label{sec:results}
Table~\ref{tab:main1} shows the trigger classification F1 scores of vanilla, KeyCP, and KeyCP++ for different LLMs.
KeyCP consistently outperforms Vanilla prompting by 5\%-30\%.
The performance of vanilla prompting varies across different models due to the different preference alignments they received.
LLaMA2 gets a very low F1 score with vanilla prompting because it shows a strong tendency for over-interpretation and results in extremely low precision. However, when equipped with KeyCP, LLaMA2 corrects its trigger profile and showcases comparable performance with other models.
KeyCP++ further achieves a consistent improvement ranging from 1\% to 8\% over KeyCP and the cumulative gain comes to 37\% at most.
This underscores the necessity of incorporating rationales into the prompt.

In Table~\ref{tab:main2} shows that KeyCP++ consistently outperforms previous fine-tuning and in-context learning baselines. Despite also utilizing the powerful DeepSeek model, ChatGPT and CodeUIE obtain performance only comparable to DEGREE. In contrast, KeyCP++ demonstrates a clear advantage, with gains ranging from 7.1\% to 12.8\%, highlighting the critical role of effective prompting. We further analyze performance across different event types, with detailed results provided in the Appendix~\ref{apsec:type-detail}.

\section{Analysis}

\begin{figure}[t]
    \centering
    \includegraphics[width=0.95\linewidth]{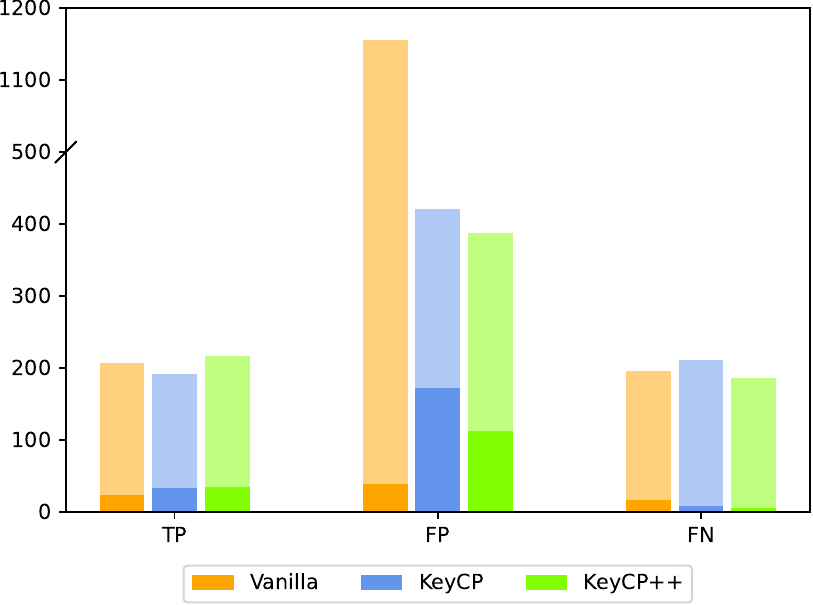}
    \caption{Number of keyword and non-keyword predictions generated by GPT3.5 on ACE2005. Shallow colors represent non-keyword predictions and deep colors represent predictions belonging to the keyword set.}
    \label{fig:kw-in-pred}
\end{figure}

\begin{table*}[t]
    \centering
    \caption{Few-shot performance on ACE05-E with LLaMA2-13B. We report precision, recall, and F1 score.}
    \begin{tabular}{cccccccccc}
        \toprule
        \multirow{2}{*}{Method}&\multicolumn{3}{c}{2-shot}&\multicolumn{3}{c}{3-shot}&\multicolumn{3}{c}{4-shot}\\
        &P&R&F&P&R&F&P&R&F\\
        \hline
        Vanilla & 5.4 & 60.8 & 9.9 & 5.5 & 65.5 & 10.1 & 5.2 & 65.3 & 9.6\\
        KeyCP & 32.7 & 48.4 & 39.0 & 31.7 & 52.6 & 39.6 & 31.5 & 53.1 & 39.5\\
        KeyCP++ & 44.5 & 46.8 & 45.6 & 43.1 & 52.6 & 47.4 & 43.8 & 53.5 & 48.2 \\
        \bottomrule
    \end{tabular}
    \label{tab:few-shot}
\end{table*}

\setlength\tabcolsep{8pt}
\begin{table}[t]
    \centering
    \caption{Ablation performance on ACE05-E with LLaMA2-13B. We report trigger classification precision, recall, and F1 score.}
    \begin{tabular}{lccc}
        \toprule
        \multirow{2}{*}{Variant}&\multicolumn{3}{c}{ACE05-E}\\
        &P&R&F\\
        \hline
        KeyCP++&44.5 &45.7 &\textbf{45.1}\\
        - Judgment&26.0&53.5&34.8\\
        - Proposal&43.5&43.9&43.7\\
        - Probing&41.8&43.9&42.9\\
        - Negative sampling&39.5&37.2&38.3\\
        - Keywords & 15.6&53.8&24.2\\
        KeyCP & 40.8 & 36.7 & 38.6 \\
        - Keyword prompting & 32.5& 36.4 & 34.3 \\
        - Keyword detection & 9.5 & 62.3 & 16.5 \\
        \bottomrule
    \end{tabular}
    \label{tab:ablation}
\end{table}

\subsection{Effect of Keywords}
Introducing keywords raises natural questions of whether the keyword dominates the trigger identification and how the keywords affect the LLMs' predictions. Therefore, we count the true positive (TP), false positive (FP), and false negative (FN) of keyword prediction and non-keyword prediction respectively. Figure~\ref{fig:kw-in-pred} shows that for both KeyCP and KeyCP++, most predictions do not belong to keywords, especially true positives. Keywords play only an auxiliary role and the main ability of event detection stems from LLMs.

The salient effect of introducing keywords is the reduction of false positives. The FP of vanilla prompting is more than twice that of KeyCP and KeyCP++, validating our hypothesis that LLMs tend to make over-interpretations without proper prompting.

Figure~\ref{fig:kw-in-pred} also reveals some side effects of KeyCP. It causes a slight decrease in total TP, a large increase in FP of keyword prediction, and a slight increase in total FN. These issues are all addressed by KeyCP++. KeyCP++ achieves a higher number of TP and reduces keyword FP, reaching the highest total TP and lowest total FP and FN.
We present case studies in Table \ref{tab:case_study} to show what false positives are eliminated by KeyCP and KeyCP++.

\subsection{Ablation Study}
We have demonstrated the effectiveness of KeyCP and KeyCP++. However, the impact of each element remains unclear. Thus, we conduct ablation experiments with the following KeyCP++ variants:
\begin{itemize}
    \item \textbf{No judgment}: Only make proposals and perform negative sampling without judgment generation. No judgment prompting.
    \item \textbf{No proposal}: Only use the detected keywords as the trigger candidates. No proposal prompting.
    \item \textbf{No probing}: Generate judgment without trigger candidates (both keywords and proposal). No proposal prompting (keyword detection is unchanged).
    \item \textbf{No negative sampling}. Uniformly sample negative samples from the other event types.
    \item \textbf{No keywords}: Apply rationale enhancement to the vanilla prompting. Remove keywords in the event description and remove keyword detection.
\end{itemize}

KeyCP variants are:
\begin{itemize}
    \item \textbf{No keyword prompting}: Remove keywords from the event description.
    \item \textbf{No keyword detection}: Remove string-matching keyword detection from prompting.
\end{itemize}

Table~\ref{tab:ablation} shows removing any element in KeyCP++ will lead to a performance drop. Without KeyCP, rationale enhancements can still obtain remarkable gains over vanilla prompting, but the performance degrades greatly. Judgment is the most critical element in the rationale-enhancement framework. We observe a serious precision decline when removing judgment, and the F1 score is even lower than KeyCP. It makes sense since proposals will encourage LLMs to make broad and divergent predictions and result in higher recall but lower precision. From another perspective, it validates that proposals can promote trigger exploration.

Removing negative sampling also causes a significant performance drop. Its performance is close to KeyCP because uniform sampling will collect a set of negative examples without any candidates. The so-constructed demonstration struggles with guiding the LLM to make proposals, and the LLM cannot generate discriminative judgment during rationale generation.

For KeyCP, removing any element will harm the precision, especially keyword detection. We believe LLMs' tendency toward over-interpretation is hard to correct by the instruction unless leveraging external assistance, such as string-matching.

\subsection{Few-shot Per Type Performance}
While this paper primarily focuses on the one-shot setting where each event type has only one positive example, we also investigate the performance of KeyCP and KeyCP++ in few-shot settings.  Given that some rare event types in ACE2005 have only a few training samples, we conduct experiments with up to 4-shot settings.

Results are presented in Table~\ref{tab:few-shot}. We find that KeyCP and KeyCP++ remain superior to vanilla prompting in few-shot settings. Notably, the advantage of KeyCP++ becomes more pronounced as the number of shots increases. The performance of Vanilla and KeyCP grows mildly and even drops in 4-shot tests because their prompting structure renders limited ability to learn from examples. In contrast, KeyCP++ can exploit informative negative examples from the enlarged training set and learn more accurate detection rules by rationale generation.

\begin{figure}[t]
    \centering
    \includegraphics[width=0.95\linewidth]{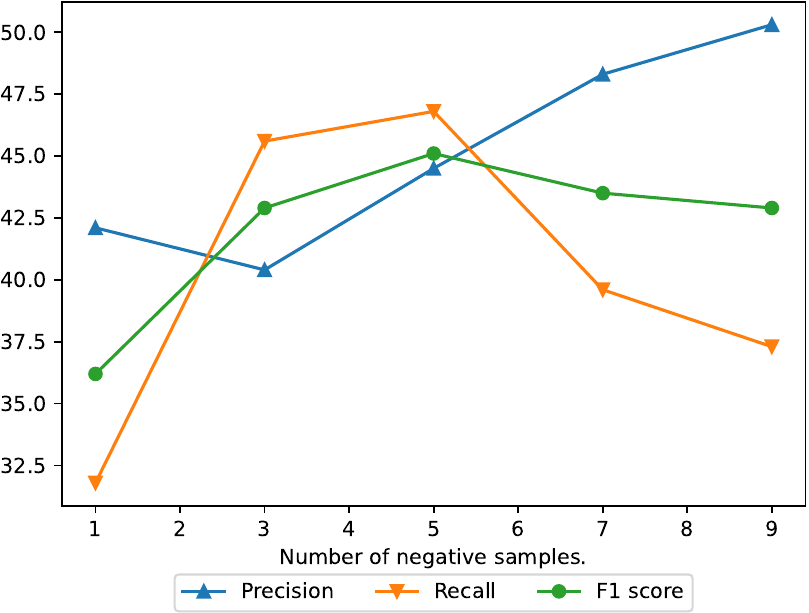}
    \caption{Performance of KeyCP++ using LLaMA2-13B on ACE-05 with varying number of negative examples.}
    \label{fig:negative}
\end{figure}

\begin{table*}[t]
    \centering
    \small
    \caption{Case studies for vanilla, KeyCP and KeyCP++ prompting.}\label{tab:case_study}
    \begin{tabular}{p{0.95\linewidth}}
         \toprule
         \textbf{Case 1} \\
         \hline
         \textbf{Query type:} Business.Start-Org\\
         \hline
         \textbf{Query:} Davies is leaving to become chairman of the London School of Economics , one of the best - known parts of the University of London.\\
         \hline
         \textbf{Keywords:} creation, establishment, find, form, create.\\
         \hline
         \textbf{Vanilla:} Based on the provided text, the trigger word related to Business.Start-Org event is leaving. \textbf{(wrong)}\\
         \hline
         \textbf{KeyCP:} The provided text does not mention any typical trigger words. Based on the provided text, no trigger signifying a Business.Start-Org event \textbf{(correct)}\\
         \hline
         \textbf{KeyCP++:} The provided text does not mention any typical trigger words. It does not describe the creation of a new organization. The text is focused on Davies leaving one position to take on another, and does not mention any new organizational structure or formation. Therefore, the text does not meet the criteria for a Business.Start-Org event. Based on the provided text, there is no trigger signifying a Business.Start-Org event. \textbf{(correct)}\\
         \midrule
         \textbf{Case 2} \\
         \hline
         \textbf{Query type:} Business.Start-Org\\
         \hline
         \textbf{Query:} We were so proud of forming an antiwar bloc with France and Germany -- but nations have not been forming such blocs for at least 12 years and someone in Moscow should have understood that it was impossible to avert this war.\\
         \hline
         \textbf{Keywords:} creation, establishment, find, form, create.\\
         \hline
         \textbf{Vanilla:} Based on the provided text, there is no trigger signifying a Business.Start-Org event.\textbf{(correct)}\\
         \hline
         \textbf{KeyCP:} Based on the provided text, the trigger word related to Business.Start-Org event is "forming" \textbf{(wrong)}\\
         \hline
         \textbf{KeyCP++:} The provided text mentions "forming". The text describes the creation of an anti-war bloc between France, Germany, and Moscow, which is not related to the establishment of a new organization. The text does not mention any details about the formation or establishment of a new entity, such as a company, corporation, or other type of organization. Therefore, the text does not meet the criteria for a Business.Start-Org event. Based on the provided text, there is no trigger signifying a Business.Start-Org event. \textbf{(correct)}\\
         \midrule
         \textbf{Case 3} \\
         \hline
         \textbf{Query type:} Life.Marry\\
         \hline
         \textbf{Query:} Anne - Marie sued Crichton for divorce in September after their marriage broke down.\\
         \hline
         \textbf{Keywords:} nuptials, marry, ceremony, wed.\\
         \hline
         \textbf{Vanilla:} Based on the provided text, the trigger word related to Life.Marry event is divorce. \textbf{(wrong)}\\
         \hline
         \textbf{KeyCP:} The provided text does not mention any typical trigger words. Based on the provided text, the trigger word for Life.Marry event is divorce. \textbf{(wrong)}\\
         \hline
         \textbf{KeyCP++:} The provided text does not mention any typical trigger words. It mentions "marriage" and "divorce", which are related to a Life.Marry event, but the text does not describe an official ceremony or legal definition of marriage. Instead, it mentions the breakdown of a marriage and the subsequent decision to seek a divorce. Based on the provided text, the trigger word related to Life.Marry event is "divorce". \textbf{(wrong)}\\
         \bottomrule
    \end{tabular}
\end{table*}

\subsection{Effect of Negative Examples}
In KeyCP and KeyCP++ prompting, the demonstration consists of one positive example and $S$ negative examples. Given that negative examples are relatively easy to acquire, we can adjust the negative example number as needed. Therefore, we conduct experiments to study the effect of varying the number of negative examples.

Table~\ref{fig:negative} shows the performance of KeyCP++ for varying numbers of negative examples. We find the best performance occurs at $S=5$. Increasing the number of negative examples beyond this point causes the LLMs to become over-conservative, thereby lowering recall.

\subsection{Case Study}
In Table~\ref{tab:case_study}, we present 3 failed cases for vanilla, KeyCP, and KeyCP++, respectively. In case 1, vanilla misidentified "leaving" as a trigger of Business.Start-Org. We speculate that the LLM reckons London School of Economics as the newly-started organization and the trigger action is the job assignment of Davies. In case 2, KeyCP over-trusted the keyword "forming" and ignored that the formed object is irrelevant to Business while KeyCP++ notes this point by actively analyzing the input text. In case 3, all methods failed because "divorce" is strongly related to the concept Marry though in the opposite direction. Even the rationale is insufficient to correct this bias.

\section{Conclusion}
In this paper, we study the in-context learning for one-shot event detection.
We find that standard ICL with input-output pairs fails to effectively align LLMs with the intricacies of the event detection task.
To this end, we decide to introduce chain-of-thought reasoning to address the weaknesses of conventional ICL and propose KeyCP and KeyCP++.
KeyCP incorporates trigger-like keywords into the event description and uses test-time keyword detection to mitigate over-interpretation.
Built on KeyCP, KeyCP++ introduces the first chain-of-thought framework tailored specifically for event detection to address the drawbacks of KeyCP.
It encourages LLMs to explore non-keyword triggers and improve the trigger identification ability by prompting with a proposal-judgment procedure.
Importantly, the rationale annotations for in-context examples are automatically generated by the LLM, eliminating the need for human annotation.
Extensive experimental results on ACE2005 and Wikievents demonstrate that keyword-centric chain-of-thought is beneficial for the event detection task, with KeyCP++ significantly outperforming previous in-context learning and supervised fine-tuning approaches.

\section{Future Work}
Although KeyCP++ has shown great effectiveness in event detection, how to generalize KeyCP++ to event argument extraction and broader information extraction applications is still under study. We are working to incorporate KeyCP++ into the code format prompting schema to achieve a unified extraction methodology.

Another problem is that, though equipped with negative sampling, the current prompting framework in KeyCP++ does not explicitly consider the discrimination between different event types, which may cause misidentification for certain types. For instance, "\textit{Conflict.Attack}" and "\textit{Life.Injure}" are easily confounded.
We expect to address this limitation by prompting LLMs to actively contrast the targeted event type with related types and corresponding examples when generating rationale.



\bibliography{mybibfile}

\begin{thebibliography}{39}
\providecommand{\natexlab}[1]{#1}
\providecommand{\url}[1]{\texttt{#1}}
\expandafter\ifx\csname urlstyle\endcsname\relax
  \providecommand{\doi}[1]{doi: #1}\else
  \providecommand{\doi}{doi: \begingroup \urlstyle{rm}\Url}\fi

\bibitem[Bird(2006)]{bird-2006-nltk}
S.~Bird.
\newblock {NLTK}: The {N}atural {L}anguage {T}oolkit.
\newblock In J.~Curran, editor, \emph{Proceedings of the {COLING}/{ACL} 2006 Interactive Presentation Sessions}, Sydney, Australia, July 2006. Association for Computational Linguistics.

\bibitem[Brown et~al.(2020)Brown, Mann, Ryder, Subbiah, Kaplan, Dhariwal, Neelakantan, Shyam, Sastry, Askell, Agarwal, Herbert-Voss, Krueger, Henighan, Child, Ramesh, Ziegler, Wu, Winter, Hesse, Chen, Sigler, Litwin, Gray, Chess, Clark, Berner, McCandlish, Radford, Sutskever, and Amodei]{brown_language_2020}
T.~Brown, B.~Mann, N.~Ryder, M.~Subbiah, J.~D. Kaplan, P.~Dhariwal, A.~Neelakantan, P.~Shyam, G.~Sastry, A.~Askell, S.~Agarwal, A.~Herbert-Voss, G.~Krueger, T.~Henighan, R.~Child, A.~Ramesh, D.~Ziegler, J.~Wu, C.~Winter, C.~Hesse, M.~Chen, E.~Sigler, M.~Litwin, S.~Gray, B.~Chess, J.~Clark, C.~Berner, S.~McCandlish, A.~Radford, I.~Sutskever, and D.~Amodei.
\newblock Language {Models} are {Few}-{Shot} {Learners}.
\newblock In \emph{Advances in {Neural} {Information} {Processing} {Systems}}. Curran Associates, Inc., 2020.

\bibitem[Chen et~al.(2024)Chen, Qin, Jiang, and Choi]{chen2024large}
R.~Chen, C.~Qin, W.~Jiang, and D.~Choi.
\newblock Is a large language model a good annotator for event extraction?
\newblock In \emph{Proceedings of the AAAI Conference on Artificial Intelligence}, volume~38, pages 17772--17780, 2024.

\bibitem[DeepSeek-AI(2025)]{deepseekai2025deepseekr1incentivizingreasoningcapability}
DeepSeek-AI.
\newblock Deepseek-r1: Incentivizing reasoning capability in llms via reinforcement learning, 2025.

\bibitem[Doddington et~al.(2004)Doddington, Mitchell, Przybocki, Ramshaw, Strassel, and Weischedel]{ace2005}
G.~R. Doddington, A.~Mitchell, M.~A. Przybocki, L.~A. Ramshaw, S.~M. Strassel, and R.~M. Weischedel.
\newblock The automatic content extraction {(ACE)} program - tasks, data, and evaluation.
\newblock In \emph{Proceedings of the Fourth International Conference on Language Resources and Evaluation, {LREC} 2004, May 26-28, 2004, Lisbon, Portugal}. European Language Resources Association, 2004.

\bibitem[Du and Cardie(2021)]{du_event_2021}
X.~Du and C.~Cardie.
\newblock Event {Extraction} by {Answering} ({Almost}) {Natural} {Questions}, Feb. 2021.
\newblock arXiv:2004.13625 [cs].

\bibitem[Gao et~al.(2023)Gao, Zhao, Yu, and Xu]{gao_exploring_2023}
J.~Gao, H.~Zhao, C.~Yu, and R.~Xu.
\newblock Exploring the {Feasibility} of {ChatGPT} for {Event} {Extraction}, Mar. 2023.
\newblock arXiv:2303.03836 [cs] version: 2.

\bibitem[Guo et~al.(2024)Guo, Li, Jin, Liu, Zeng, Liu, Li, Yang, Bai, Guo, et~al.]{guo_retrieval-augmented_2024}
Y.~Guo, Z.~Li, X.~Jin, Y.~Liu, Y.~Zeng, W.~Liu, X.~Li, P.~Yang, L.~Bai, J.~Guo, et~al.
\newblock Retrieval-augmented code generation for universal information extraction.
\newblock In \emph{CCF International Conference on Natural Language Processing and Chinese Computing}. Springer, 2024.

\bibitem[Han et~al.(2021)Han, Hsu, Sun, Baylon, Ning, Roth, and Peng]{han2021ester}
R.~Han, I.-H. Hsu, J.~Sun, J.~Baylon, Q.~Ning, D.~Roth, and N.~Peng.
\newblock Ester: A machine reading comprehension dataset for reasoning about event semantic relations.
\newblock In \emph{Proceedings of the 2021 Conference on Empirical Methods in Natural Language Processing}, 2021.

\bibitem[Hsu et~al.(2022)Hsu, Huang, Boschee, Miller, Natarajan, Chang, and Peng]{hsu_degree_2022}
I.-H. Hsu, K.-H. Huang, E.~Boschee, S.~Miller, P.~Natarajan, K.-W. Chang, and N.~Peng.
\newblock {DEGREE}: {A} {Data}-{Efficient} {Generation}-{Based} {Event} {Extraction} {Model}.
\newblock In \emph{NAACL: {Human} {Language} {Technologies}}, Seattle, United States, July 2022. Association for Computational Linguistics.

\bibitem[Huang et~al.(2023)Huang, Xu, Zeng, Chen, Yang, and E]{huang_iteratively_2023}
G.~Huang, R.~Xu, Y.~Zeng, J.~Chen, Z.~Yang, and W.~E.
\newblock An {Iteratively} {Parallel} {Generation} {Method} with the {Pre}-{Filling} {Strategy} for {Document}-level {Event} {Extraction}.
\newblock In H.~Bouamor, J.~Pino, and K.~Bali, editors, \emph{EMNLP}, Singapore, Dec. 2023. Association for Computational Linguistics.

\bibitem[Jiang et~al.(2023)Jiang, Sablayrolles, Mensch, Bamford, Chaplot, Casas, Bressand, Lengyel, Lample, Saulnier, Lavaud, Lachaux, Stock, Scao, Lavril, Wang, Lacroix, and Sayed]{jiang_mistral_2023}
A.~Q. Jiang, A.~Sablayrolles, A.~Mensch, C.~Bamford, D.~S. Chaplot, D.~d.~l. Casas, F.~Bressand, G.~Lengyel, G.~Lample, L.~Saulnier, L.~R. Lavaud, M.-A. Lachaux, P.~Stock, T.~L. Scao, T.~Lavril, T.~Wang, T.~Lacroix, and W.~E. Sayed.
\newblock Mistral {7B}, Oct. 2023.
\newblock arXiv:2310.06825 [cs].

\bibitem[Kojima et~al.(2022)Kojima, Gu, Reid, Matsuo, and Iwasawa]{kojima_large_2022}
T.~Kojima, S.~S. Gu, M.~Reid, Y.~Matsuo, and Y.~Iwasawa.
\newblock Large {Language} {Models} are {Zero}-{Shot} {Reasoners}.
\newblock \emph{Advances in Neural Information Processing Systems}, Dec. 2022.

\bibitem[Li et~al.(2020{\natexlab{a}})Li, Peng, Chen, Wang, Pan, Lyu, and Zhu]{li_event_2020}
F.~Li, W.~Peng, Y.~Chen, Q.~Wang, L.~Pan, Y.~Lyu, and Y.~Zhu.
\newblock Event {Extraction} as {Multi}-turn {Question} {Answering}.
\newblock In T.~Cohn, Y.~He, and Y.~Liu, editors, \emph{Findings of the {Association} for {Computational} {Linguistics}: {EMNLP} 2020}, Online, Nov. 2020{\natexlab{a}}. Association for Computational Linguistics.

\bibitem[Li et~al.(2020{\natexlab{b}})Li, Zareian, Lin, Pan, Whitehead, Chen, Wu, Ji, Chang, Voss, Napierski, and Freedman]{li-etal-2020-gaia}
M.~Li, A.~Zareian, Y.~Lin, X.~Pan, S.~Whitehead, B.~Chen, B.~Wu, H.~Ji, S.-F. Chang, C.~Voss, D.~Napierski, and M.~Freedman.
\newblock {GAIA}: A fine-grained multimedia knowledge extraction system.
\newblock In A.~Celikyilmaz and T.-H. Wen, editors, \emph{Proceedings of the 58th Annual Meeting of the Association for Computational Linguistics: System Demonstrations}, Online, July 2020{\natexlab{b}}. Association for Computational Linguistics.

\bibitem[Li et~al.(2021)Li, Ji, and Han]{li_document-level_2021}
S.~Li, H.~Ji, and J.~Han.
\newblock Document-{Level} {Event} {Argument} {Extraction} by {Conditional} {Generation}.
\newblock In K.~Toutanova, A.~Rumshisky, L.~Zettlemoyer, D.~Hakkani-Tur, I.~Beltagy, S.~Bethard, R.~Cotterell, T.~Chakraborty, and Y.~Zhou, editors, \emph{Proceedings of the 2021 {Conference} of the {North} {American} {Chapter} of the {Association} for {Computational} {Linguistics}: {Human} {Language} {Technologies}}, Online, June 2021. Association for Computational Linguistics.

\bibitem[Lin et~al.(2020)Lin, Ji, Huang, and Wu]{lin_joint_2020}
Y.~Lin, H.~Ji, F.~Huang, and L.~Wu.
\newblock A {Joint} {Neural} {Model} for {Information} {Extraction} with {Global} {Features}.
\newblock In D.~Jurafsky, J.~Chai, N.~Schluter, and J.~Tetreault, editors, \emph{ACL}, Online, July 2020. Association for Computational Linguistics.

\bibitem[Liu et~al.(2020)Liu, Chen, Liu, Bi, and Liu]{liu_event_2020}
J.~Liu, Y.~Chen, K.~Liu, W.~Bi, and X.~Liu.
\newblock Event {Extraction} as {Machine} {Reading} {Comprehension}.
\newblock In B.~Webber, T.~Cohn, Y.~He, and Y.~Liu, editors, \emph{{EMNLP}}, Online, Nov. 2020. Association for Computational Linguistics.

\bibitem[Liu et~al.(2022)Liu, Huang, Shi, and Wang]{liu_dynamic_2022}
X.~Liu, H.~Huang, G.~Shi, and B.~Wang.
\newblock Dynamic {Prefix}-{Tuning} for {Generative} {Template}-based {Event} {Extraction}, May 2022.
\newblock arXiv:2205.06166 [cs].

\bibitem[Lu et~al.(2021)Lu, Lin, Xu, Han, Tang, Li, Sun, Liao, and Chen]{lu_text2event_2021}
Y.~Lu, H.~Lin, J.~Xu, X.~Han, J.~Tang, A.~Li, L.~Sun, M.~Liao, and S.~Chen.
\newblock {Text2Event}: {Controllable} {Sequence}-to-{Structure} {Generation} for {End}-to-end {Event} {Extraction}.
\newblock In \emph{ACL}, Online, Aug. 2021. Association for Computational Linguistics.

\bibitem[Lu et~al.(2022)Lu, Liu, Dai, Xiao, Lin, Han, Sun, and Wu]{lu_unified_2022}
Y.~Lu, Q.~Liu, D.~Dai, X.~Xiao, H.~Lin, X.~Han, L.~Sun, and H.~Wu.
\newblock Unified {Structure} {Generation} for {Universal} {Information} {Extraction}.
\newblock In S.~Muresan, P.~Nakov, and A.~Villavicencio, editors, \emph{ACL}, Dublin, Ireland, May 2022. Association for Computational Linguistics.

\bibitem[Ma et~al.(2025)Ma, Zhao, Zhang, He, and Kong]{ma_non-myopic_2025}
C.~Ma, H.~Zhao, J.~Zhang, J.~He, and L.~Kong.
\newblock Non-myopic {Generation} of {Language} {Models} for {Reasoning} and {Planning}.
\newblock 2025.

\bibitem[Nguyen et~al.(2016)Nguyen, Cho, and Grishman]{nguyen_joint_2016}
T.~H. Nguyen, K.~Cho, and R.~Grishman.
\newblock Joint {Event} {Extraction} via {Recurrent} {Neural} {Networks}.
\newblock In K.~Knight, A.~Nenkova, and O.~Rambow, editors, \emph{NAACL: {Human} {Language} {Technologies}}, San Diego, California, June 2016. Association for Computational Linguistics.

\bibitem[OpenAI(2023)]{openai_gpt-4_2023}
OpenAI.
\newblock {GPT}-4 {Technical} {Report}, Dec. 2023.
\newblock arXiv:2303.08774 [cs].

\bibitem[Pang et~al.(2023)Pang, Cao, Ding, and Luo]{pang_guideline_2023}
C.~Pang, Y.~Cao, Q.~Ding, and P.~Luo.
\newblock Guideline {Learning} for {In}-{Context} {Information} {Extraction}.
\newblock In H.~Bouamor, J.~Pino, and K.~Bali, editors, \emph{EMNLP}, Singapore, Dec. 2023. Association for Computational Linguistics.

\bibitem[Pouran Ben~Veyseh et~al.(2021)Pouran Ben~Veyseh, Lai, Dernoncourt, and Nguyen]{pouran_ben_veyseh_unleash_2021}
A.~Pouran Ben~Veyseh, V.~Lai, F.~Dernoncourt, and T.~H. Nguyen.
\newblock Unleash {GPT}-2 {Power} for {Event} {Detection}.
\newblock In C.~Zong, F.~Xia, W.~Li, and R.~Navigli, editors, \emph{ACL}, Online, Aug. 2021. Association for Computational Linguistics.

\bibitem[Touvron et~al.(2023)Touvron, Lavril, Izacard, Martinet, Lachaux, Lacroix, Rozière, Goyal, Hambro, Azhar, Rodriguez, Joulin, Grave, and Lample]{touvron_llama_2023}
H.~Touvron, T.~Lavril, G.~Izacard, X.~Martinet, M.-A. Lachaux, T.~Lacroix, B.~Rozière, N.~Goyal, E.~Hambro, F.~Azhar, A.~Rodriguez, A.~Joulin, E.~Grave, and G.~Lample.
\newblock {LLaMA}: {Open} and {Efficient} {Foundation} {Language} {Models}, Feb. 2023.
\newblock arXiv:2302.13971 [cs].

\bibitem[Wadden et~al.(2019)Wadden, Wennberg, Luan, and Hajishirzi]{wadden_entity_2019}
D.~Wadden, U.~Wennberg, Y.~Luan, and H.~Hajishirzi.
\newblock Entity, {Relation}, and {Event} {Extraction} with {Contextualized} {Span} {Representations}.
\newblock In K.~Inui, J.~Jiang, V.~Ng, and X.~Wan, editors, \emph{{EMNLP}-{IJCNLP}}, Hong Kong, China, Nov. 2019. Association for Computational Linguistics.

\bibitem[Wang et~al.(2023)Wang, Li, and Ji]{wang_code4struct_2023}
X.~Wang, S.~Li, and H.~Ji.
\newblock {Code4Struct}: {Code} {Generation} for {Few}-{Shot} {Event} {Structure} {Prediction}.
\newblock In A.~Rogers, J.~Boyd-Graber, and N.~Okazaki, editors, \emph{Proceedings of the 61st {Annual} {Meeting} of the {Association} for {Computational} {Linguistics} ({Volume} 1: {Long} {Papers})}, Toronto, Canada, July 2023. Association for Computational Linguistics.

\bibitem[Wang et~al.(2024)Wang, Zhao, Wang, Huang, Fan, Zhang, Wang, Wang, and Liu]{wang_strategic_2024-1}
Y.~Wang, S.~Zhao, Z.~Wang, H.~Huang, M.~Fan, Y.~Zhang, Z.~Wang, H.~Wang, and T.~Liu.
\newblock Strategic {Chain}-of-{Thought}: {Guiding} {Accurate} {Reasoning} in {LLMs} through {Strategy} {Elicitation}, 2024.
\newblock arXiv:2409.03271 [cs].

\bibitem[Wei et~al.(2022)Wei, Wang, Schuurmans, Bosma, Ichter, Xia, Chi, Le, and Zhou]{wei_chain--thought_2022}
J.~Wei, X.~Wang, D.~Schuurmans, M.~Bosma, B.~Ichter, F.~Xia, E.~Chi, Q.~V. Le, and D.~Zhou.
\newblock Chain-of-{Thought} {Prompting} {Elicits} {Reasoning} in {Large} {Language} {Models}.
\newblock \emph{Advances in Neural Information Processing Systems}, Dec. 2022.

\bibitem[Wei et~al.(2023)Wei, Cui, Cheng, Wang, Zhang, Huang, Xie, Xu, Chen, Zhang, Jiang, and Han]{wei_zero-shot_2023}
X.~Wei, X.~Cui, N.~Cheng, X.~Wang, X.~Zhang, S.~Huang, P.~Xie, J.~Xu, Y.~Chen, M.~Zhang, Y.~Jiang, and W.~Han.
\newblock Zero-{Shot} {Information} {Extraction} via {Chatting} with {ChatGPT}, Feb. 2023.
\newblock arXiv:2302.10205 [cs].

\bibitem[Wu et~al.(2024)Wu, Sun, Li, Welleck, and Yang]{wu_inference_2024}
Y.~Wu, Z.~Sun, S.~Li, S.~Welleck, and Y.~Yang.
\newblock Inference {Scaling} {Laws}: {An} {Empirical} {Analysis} of {Compute}-{Optimal} {Inference} for {Problem}-{Solving} with {Language} {Models}, 2024.
\newblock arXiv:2408.00724 [cs].

\bibitem[Xia et~al.(2023)Xia, Yu, Wang, Xuan, and Luo]{dafs}
N.~Xia, H.~Yu, Y.~Wang, J.~Xuan, and X.~Luo.
\newblock {DAFS:} a domain aware few shot generative model for event detection.
\newblock \emph{Mach. Learn.}, \penalty0 (3), 2023.

\bibitem[Xu et~al.()Xu, Chen, Peng, Zhang, Xu, Zhao, Wu, Zheng, Wang, and Chen]{xu_large_2024}
D.~Xu, W.~Chen, W.~Peng, C.~Zhang, T.~Xu, X.~Zhao, X.~Wu, Y.~Zheng, Y.~Wang, and E.~Chen.
\newblock Large language models for generative information extraction: a survey.

\bibitem[Yang et~al.(2019)Yang, Feng, Qiao, Kan, and Li]{yang_exploring_2019}
S.~Yang, D.~Feng, L.~Qiao, Z.~Kan, and D.~Li.
\newblock Exploring {Pre}-trained {Language} {Models} for {Event} {Extraction} and {Generation}.
\newblock In A.~Korhonen, D.~Traum, and L.~Màrquez, editors, \emph{ACL}, Florence, Italy, July 2019. Association for Computational Linguistics.

\bibitem[Zhang et~al.(2020)Zhang, Liu, Pan, Song, and Leung]{zhang2020aser}
H.~Zhang, X.~Liu, H.~Pan, Y.~Song, and C.~W.-K. Leung.
\newblock Aser: A large-scale eventuality knowledge graph.
\newblock In \emph{Proceedings of the web conference 2020}, 2020.

\bibitem[Zhang et~al.(2025)Zhang, Lyu, Sun, Wang, Zhang, Guo, Wang, Muennighoff, King, Liu, and Ma]{zhang_what_2025}
Q.~Zhang, F.~Lyu, Z.~Sun, L.~Wang, W.~Zhang, Z.~Guo, Y.~Wang, N.~Muennighoff, I.~King, X.~Liu, and C.~Ma.
\newblock What, {How}, {Where}, and {How} {Well}? {A} {Survey} on {Test}-{Time} {Scaling} in {Large} {Language} {Models}, 2025.
\newblock arXiv:2503.24235 [cs].

\bibitem[Zhao et~al.(2023)Zhao, Gong, Yang, Dong, Lu, and Li]{zhao_demosg_2023}
G.~Zhao, X.~Gong, X.~Yang, G.~Dong, S.~Lu, and S.~Li.
\newblock {DemoSG}: {Demonstration}-enhanced {Schema}-guided {Generation} for {Low}-resource {Event} {Extraction}.
\newblock In H.~Bouamor, J.~Pino, and K.~Bali, editors, \emph{Findings of the {Association} for {Computational} {Linguistics}: {EMNLP} 2023}, Singapore, Dec. 2023. Association for Computational Linguistics.

\end{thebibliography}

\appendix

\section{Performance Across Event Types}\label{apsec:type-detail}
\begin{figure*}[t]
    \centering
    \includegraphics[width=0.98\linewidth]{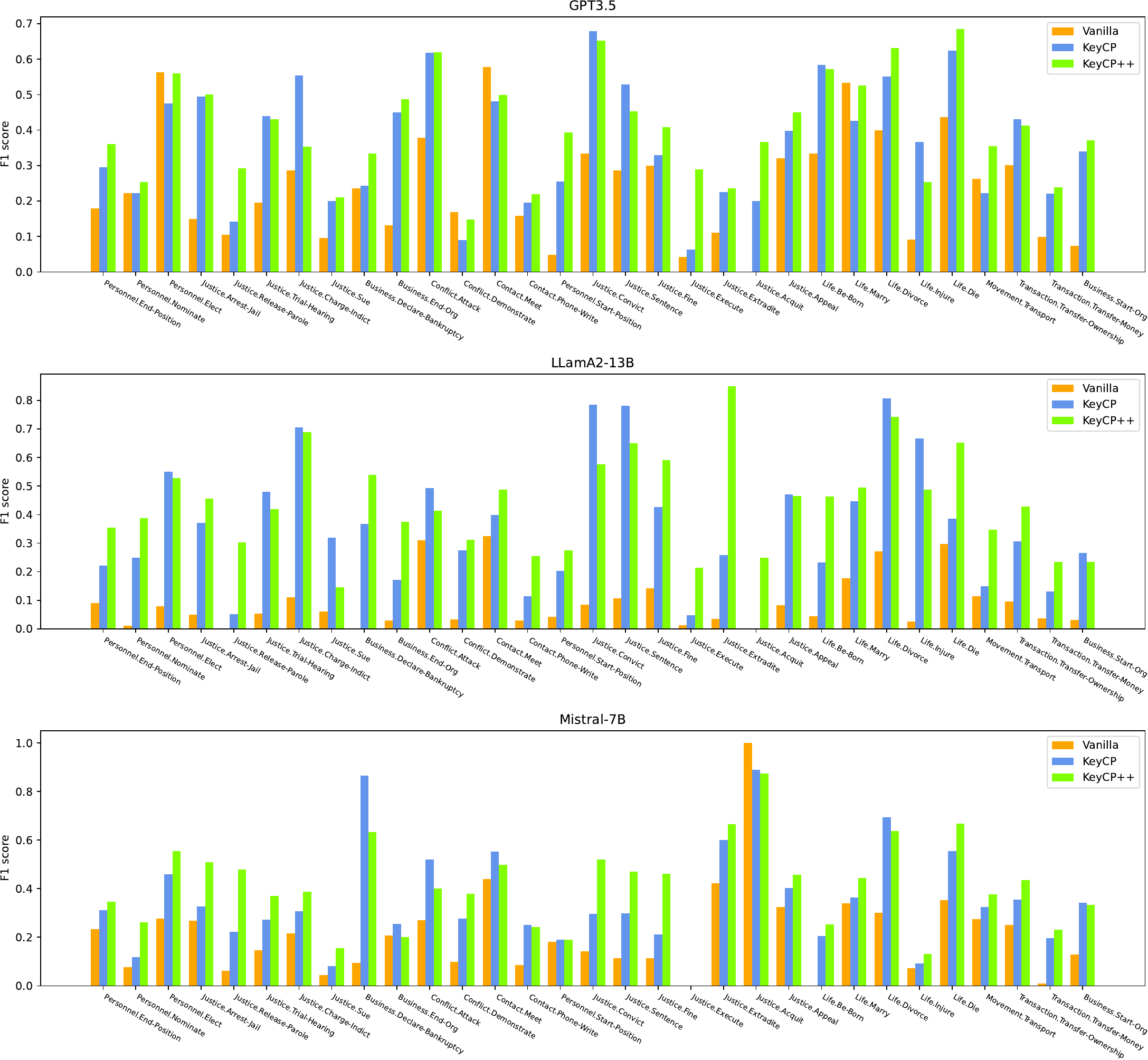}
    \caption{Performance across event types on ACE2005.}
    \label{fig:ace2005-detail}
\end{figure*}

We report the F1 scores across all event types appearing in the ACE2005 test set in Figure~\ref{fig:ace2005-detail}. We can find the improvements brought by KeyCP and KeyCP++ are consistent in general. The performance varies across different types and across different models, depending on their preference alignment.

\section{Keywords Generation}\label{apsec:keywords}
First, we generate candidate keywords using GPT3.5 with the prompt shown in Table~\ref{tab:gen_keywords}. We repeat generation five times for each event type and take candidates appearing more than three times. For ACE2005, we add a few keywords used in DEGREE~\citep{hsu_degree_2022} as examples. Subsequently, we prompt GPT3.5 to check if each candidate is related to the corresponding event with the prompt shown in Table~\ref{tab:gen_keywords}.

Table~\ref{tab:ace2005_keywords} and Table~\ref{tab:wikievents_keywords} show the generated keywords for ACE2005 and WikiEvents respectively. Although we have leverage voting in the candidate generation, there are still many low-quality generations. Nevertheless, KeyCP and KeyCP++ can still work well.
\begin{table*}
    \centering
    \caption{Prompt used to generate keywords. Here we take Transaction.Transfer-Money event as an example.}\label{tab:gen_keywords}
    \begin{tabular}{p{\linewidth}}
         \toprule
         Here is the definition of event \textit{Transaction.Transfer-Money}:\\
         TRANSFER-MONEY Events refer to the giving, receiving, borrowing, or lending money when it is not in the context of purchasing something. The canonical examples are: (1) people giving money to organizations (and getting nothing tangible in return); and (2) organizations lending money to people or other orgs.\\
         Please find more trigger words (verbs, nouns, adjectives or adverbs) that can signify a \textit{Transaction.Transfer-Money} event happens from the definition. Each trigger word should be only one word. Please only output those you are confident in. The word literally appearing in the definition. The output should be JSON format like {"answer": [word1, word2, ...]}\\
         \bottomrule
    \end{tabular}
\end{table*}

\begin{table*}
    \centering
    \caption{Prompt used to check keywords. Here we take Transaction.Transfer-Money event as an example.}\label{tab:check_keywords}
    \begin{tabular}{p{\linewidth}}
         \toprule
         Here is the definition of event \textit{Transaction.Transfer-Money}:\\
         TRANSFER-MONEY Events refer to the giving, receiving, borrowing, or lending money when it is not in the context of purchasing something. The canonical examples are: (1) people giving money to organizations (and getting nothing tangible in return); and (2) organizations lending money to people or other orgs.\\
         According to the event definition, is the word "XXX" related to the event Transaction.Transfer-Money? Only answer yes or no.\\
         \bottomrule
    \end{tabular}
\end{table*}

\begin{table*}
    \centering
    \caption{Generated keywords for ACE2005.}\label{tab:ace2005_keywords}
    \begin{tabular}{l|p{0.6\linewidth}}
        \hline
         \textbf{Event type} & \textbf{Keywords} \\
         \hline
         Life.Be-Born & birth, bore  \\
         \hline
         Life.Marry & marry, ceremony, nuptials, wed \\
         \hline
         Life.Divorce & divorce, terminate, separate, split\\
         \hline
         Life.Injure & injure, wounded, hurt, painful, injury, trauma, harm\\
         \hline
         Life.Die & die, kill, cease, perish, expire\\
         \hline
         Movement.Transport & travel, go, relocate, shift, move, transport\\
         \hline
         Transaction.Transfer-Ownership & buy, sell, receive, borrow \\
         \hline
         Transaction.Transfer-Money & pay, donation, loan, give, receive, borrow \\
         \hline
         Business.Start-Org & found, create, creation, form, establishment \\
         \hline
         Business.Merge-Org & merge, venture, joint, form \\
         \hline
         Business.Declare-Bankruptcy & bankruptcy, bankrupt, debt, protection, request, balance, sheet, collection, negative \\
         \hline
         Business.End-Org & dissolve, cease \\
         \hline
         Conflict.Attack & attack, coup, clash, war, fight, harm, damage, violence, gunfire \\
         \hline
         Conflict.Demonstrate & protest, demonstrate, official, strike, area, demand, public, sit-in, riot\\
         \hline
         Contact.Meet & meet, assemble, encounter, converse, interact \\
         \hline
         Contact.Phone-Write & call, engage, discuss, dialogue, converse \\
         \hline
         Personnel.Start-Position & appoint, change, begin \\
         \hline
         Personnel.End-Position & resign, fire, leave, terminate \\
         \hline
         Personnel.Nominate & named, nominate, nomination, channel, propose \\
         \hline
         Personnel.Elect & election, elect, determine, win \\
         \hline
         Justice.Arrest-Jail & arrest, jail, imprison, detain, apprehend \\
         \hline
         Justice.Release-Parole & release, free, drop, grant \\
         \hline
         Justice.Trial-Hearing & trial, hearing, discuss, gather, initiate \\
         \hline
         Justice.Charge-Indict & charge, accuse \\
         \hline
         Justice.Sue & sue, accuse, commit, proceed \\
         \hline
         Justice.Convict & convict, guilty, conviction, prosecution \\
         \hline
         Justice.Sentence & sentence, issue, incarceration, punishment \\
         \hline
         Justice.Fine & fine, court, proceed, punishment \\
         \hline
         Justice.Execute & execution, execute, take \\
         \hline
         Justice.Extradite & extradition, extradite, send, legal, proceed \\
         \hline
         Justice.Acquit & acquittal, acquit, conviction, drop \\
         \hline
         Justice.Appeal & review, trigger \\
         \hline
         Justice.Pardon & pardon, lift, impose \\
         \hline
    \end{tabular}
\end{table*}

\begin{table*}
    \centering
    \caption{Generated keywords for WikiEvents.}\label{tab:wikievents_keywords}
    \begin{tabular}{l|p{0.4\linewidth}}
        \hline
         \textbf{Event type} & \textbf{Keywords} \\
         \hline
         ArtifactExistence.DamageDestroyDisableDismantle & disassemble, dismantle, damage, disable, defuse, destroy \\
        \hline
        ArtifactExistence.ManufactureAssemble & mix, assemble, put \\
        \hline
        Cognitive.IdentifyCategorize & identify, identity, relevant, category \\
        \hline
        Cognitive.Inspection & target, inspection, observation \\
        \hline
        Cognitive.Research & experiment, review, test \\
        \hline
        Cognitive.TeachingTrainingLearning & educate, instruct, train \\
        \hline
        Conflict.Attack & attack, violent, assault, harm, bomb \\
        \hline
        Conflict.Defeat & loss, overthrow, failure \\
        \hline
        Conflict.Demonstrate & demonstration, violent, march, riot, protest \\
        \hline
        Contact.Contact & email, communication, medium, communicate, phone \\
        \hline
        Contact.RequestCommand & ask, order \\
        \hline
        Contact.ThreatenCoerce & blackmail, intimidate, menace \\
        \hline
        Control.ImpedeInterfereWith & impede, hamper, obstruct, interfere \\
        \hline
        Disaster.Crash & crash, vehicle, collision \\
        \hline
        Disaster.DiseaseOutbreak & outbreak, region, area, disease, country \\
        \hline
        GenericCrime.GenericCrime & crime \\
        \hline
        Justice.Acquit & acquit, fail, drop \\
        \hline
        Justice.ArrestJailDetain & detain, detention \\
        \hline
        Justice.ChargeIndict & charge, indict, accuse \\
        \hline
        Justice.Convict & convict, guilty, find \\
        \hline
        Justice.InvestigateCrime & investigate, pursue, probe, detect \\
        \hline
        Justice.ReleaseParole & drop, release, grant \\
        \hline
        Justice.Sentence & sentence, punishment, incarceration \\
        \hline
        Justice.TrialHearing & hear, gather, legal, matter \\
        \hline
        Life.Die & die, death, expiration, cease, demise \\
        \hline
        Life.Infect & infect, spread, contract, transmission \\
        \hline
        Life.Injure & hurt, injure, injury, trauma \\
        \hline
        Medical.Intervention & diagnosis, medical, intervene, intervention, condition, treatment \\
        \hline
        Movement.Transportation & move, place, smuggle, movement, evacuation, traffic, transportation, border \\
        \hline
        Personnel.EndPosition & resign, leave, terminate \\
        \hline
        Personnel.StartPosition & begin \\
        \hline
        Transaction.Donation & asset, provision, material, donate, donation, aid, extension \\
        \hline
        Transaction.ExchangeBuySell & buy, sell, possession, money, acquisition, ownership, obtain, barter \\
        \hline
    \end{tabular}
\end{table*}

\section{Prompting examples}\label{apsec:example}
In Table~\ref{tab:pos_example} we present the full prompting example of KeyCP++. There is a small difference between the practical prompts and the illustration in Figure 3 in the paper. We move the keywords from the event description to the instruction in each demonstration example and the instance. This modification makes LLMs pay more attention to these keywords and perform better in our experiments.
\begin{table*}
    \centering
    \caption{A prompt example of positive prediction.}\label{tab:pos_example}
    \begin{tabular}{p{\linewidth}}
         \toprule
         \textit{Query type: Transaction.Transfer-Money}\\
         \hline
\textbf{Task instruction:} This is an event detection task where the goal is to extract the trigger word indicating a certain type event happens. If there is no such a trigger, tell me none. The definition of the target event is:\\
\hline
\textbf{Event description:} TRANSFER-MONEY Events refer to the giving, receiving, borrowing, or lending money when it is not in the context of purchasing something. The canonical examples are: (1) people giving money to organizations (and getting nothing tangible in return); and (2) organizations lending money to people or other orgs.\\
\hline
\textbf{Demonstration:} If the following query sentence does not mention a Transaction.Transfer-Money event, tell me there is no trigger word. Otherwise, please find the trigger word related to Transaction.Transfer-Money event in following text. Similar words are donation, give, loan, borrow, receive, pay.

Query: Though , of course , there 's always the chance that these fund managers were stupid and shortsighted , not actually criminal ; apparently , the money stolen was lent to or invested in companies the managers controlled -- so they 're guilty of front - running , at the least .

The provided text does not mention any typical trigger words.If we relax the criteria for trigger words, the provided text additionally mentions "lent" and "invested".The trigger word "lent" is appropriate for the Transaction.Transfer-Money event because it specifically involves the act of borrowing or lending money, which aligns with the definition of the event. On the other hand, "invested" does not directly signify the act of giving, receiving, borrowing, or lending money as described in the context of the Transaction.Transfer-Money event. Based on the provided text, the trigger word signifying a Transaction.Transfer-Money event is lent

If the following query sentence does not mention a Transaction.Transfer-Money event, tell me there is no trigger word. Otherwise, please find the trigger word related to Transaction.Transfer-Money event in following text. Similar words are donation, give, loan, borrow, receive, pay.

Query: When she 's in Germany , Lynch will be treated for bullet wounds and broken bones .

The provided text does not mention any typical trigger words.it does not involve giving, receiving, borrowing, or lending money outside the context of purchasing something. Based on the provided text, there is no trigger signifying a Transaction.Transfer-Money event

If the following query sentence does not mention a Transaction.Transfer-Money event, tell me there is no trigger word. Otherwise, please find the trigger word related to Transaction.Transfer-Money event in following text. Similar words are donation, give, loan, borrow, receive, pay.

Query: Earlier this week , security police detained the Monitor 's managing editor , Nhial Bol , for 24 hours .

The provided text does not mention any typical trigger words.it does not involve any giving, receiving, borrowing, or lending of money as described in the event description. Based on the provided text, there is no trigger signifying a Transaction.Transfer-Money event

If the following query sentence does not mention a Transaction.Transfer-Money event, tell me there is no trigger word. Otherwise, please find the trigger word related to Transaction.Transfer-Money event in following text. Similar words are donation, give, loan, borrow, receive, pay.

Query: The six nations contributing troops are Italy , Spain , Ukraine , Denmark , the Netherlands and Bulgaria , the official said . Representatives of those countries will meet with British officials May 7 and Polish officials May 22 to determine what forces each country will contribute and whether they will be put under British or Polish command .

The provided text does not mention any typical trigger words.If we relax the criteria for trigger words, the provided text additionally mentions "contribute" and "contributing".the mention of countries contributing troops does not involve the giving, receiving, borrowing, or lending of money as specified in the event description. Based on the provided text, there is no trigger signifying a Transaction.Transfer-Money event

If the following query sentence does not mention a Transaction.Transfer-Money event, tell me there is no trigger word. Otherwise, please find the trigger word related to Transaction.Transfer-Money event in following text. Similar words are donation, give, loan, borrow, receive, pay.

Query: u.s. officials believe zarqawi planned the assassination of u.s. diplomat lawrence foley in jordan last october .

The provided text does not mention any typical trigger words.it does not involve the giving, receiving, borrowing, or lending of money as described in the event description. Based on the provided text, there is no trigger signifying a Transaction.Transfer-Money event

If the following query sentence does not mention a Transaction.Transfer-Money event, tell me there is no trigger word. Otherwise, please find the trigger word related to Transaction.Transfer-Money event in following text. Similar words are donation, give, loan, borrow, receive, pay.

Query: Hmm ... I wonder how many success stories are out there from people who chose to have a civil ceremony , and went on later to be happily sealed .

The provided text does not mention any typical trigger words.it does not involve giving, receiving, borrowing, or lending money in a context that is not related to purchasing something. Based on the provided text, there is no trigger signifying a Transaction.Transfer-Money event \\

\hline
\textbf{Instance:} If the following query sentence does not mention a Transaction.Transfer-Money event, tell me there is no trigger word. Otherwise, please find the trigger word related to Transaction.Transfer-Money event in following text. Similar words are donation, give, loan, borrow, receive, pay.

Query: Kommersant business daily joined in , declaring in a furious front - page headline : " The United States is demanding that Russia , France and Germany pay for the Iraqi war . "\\
\hline
\textbf{Output:} The provided text mentions pay. The word "pay" is a trigger word for the Transaction.Transfer-Money event because it indicates that money is being transferred from one party to another. In this case, the text states that the United States is demanding that Russia, France, and Germany pay for the Iraqi war, which aligns with the definition of the Transaction.Transfer-Money event.

Based on the provided text, the trigger word signifying a Transaction.Transfer-Money event is pay.\\
         \bottomrule
    \end{tabular}
\end{table*}

\end{document}